\documentclass[final]{cvpr}

\usepackage{times}
\usepackage{epsfig}
\usepackage{graphicx}
\usepackage{amsmath}
\usepackage{amssymb}
\usepackage{booktabs}
\usepackage{multirow}
\usepackage[pagebackref=true,breaklinks=true,colorlinks,bookmarks=false]{hyperref}
\def\cvprPaperID{8145} % *** Enter the CVPR Paper ID here
\def\confYear{CVPR 2021}

\begin{document}

%%%%%%%%% TITLE
\title{Improving Accuracy of Binary Neural Networks using Unbalanced Activation Distribution}

\author{Hyungjun Kim$^1$\;\;\;\;\;\;\;\;Jihoon Park$^1$\;\;\;\;\;\;\;\;Changhun Lee$^1$\;\;\;\;\;\;\;\;Jae-Joon Kim$^{1,2}$\\
\normalsize{$^1$Department of Convergence IT Engineering, $^2$Graduate School of Artificial Intelligence}\\
\normalsize{Pohang University of Science and Technology (POSTECH), Korea}\\
{\tt\small \{hyungjun.kim, jihoon.park, changhun.lee, jaejoon\}@postech.ac.kr}
% \and
}

\maketitle

%%%%%%%%% ABSTRACT
\begin{abstract}
Binarization of neural network models is considered as one of the promising methods to deploy deep neural network models on resource-constrained environments such as mobile devices.
However, Binary Neural Networks (BNNs) tend to suffer from severe accuracy degradation compared to the full-precision counterpart model.
Several techniques were proposed to improve the accuracy of BNNs.
One of the approaches is to balance the distribution of binary activations so that the amount of information in the binary activations becomes maximum.
Based on extensive analysis, in stark contrast to previous work, we argue that unbalanced activation distribution can actually improve the accuracy of BNNs.
We also show that adjusting the threshold values of binary activation functions results in the unbalanced distribution of the binary activation, which increases the accuracy of BNN models.
Experimental results show that the accuracy of previous BNN models (e.g. XNOR-Net and Bi-Real-Net) can be improved by simply shifting the threshold values of binary activation functions without requiring any other modification.
\end{abstract}

%%%%%%%%% BODY TEXT
\section{Introduction}
Deep Neural Networks (DNNs) have achieved human-level performance in many computer vision tasks such as image classification, object detection, and segmentation.
However, the increased compute cost and memory requirement of large DNN models pose a burden on resource-constrained environments such as mobile devices.
To mitigate this problem, various techniques including network quantization~\cite{pact,bnn,abc-net,xnor,dorefa}, network pruning~\cite{deepcompression,structuredpruning}, and efficient architecture design~\cite{mobilenet,efficientnet,shift} were introduced to reduce the compute cost and memory requirement of DNN models.
Among them, the network quantization technique is being actively studied and recent works have shown that a DNN model can even be quantized to a 1-bit model~\cite{bnn,bireal,rtb,xnor}.
When a DNN model is binarized to a Binary Neural Network (BNN) model, the memory requirement of the model is reduced by 32x since 32-bit floating-point weights can be represented by 1-bit weights.
In addition, high precision multiply-and-accumulate operations can be replaced by XNOR-and-popcount logics in BNNs since both activation and weight have 1-bit precision.
Due to the lightweight nature, BNNs are garnering interests as a promising solution for DNN computing on edge devices.

However, BNNs still suffer from the accuracy degradation caused by the aggressive quantization (32-bit to 1-bit).
While recent researches have shown that a DNN model can be quantized to 2-bit precision with marginal accuracy loss~\cite{lsq}, a severe performance gap still exists between a BNN model and its full-precision counterpart DNN model.
In general, it is known that weight quantization is much easier than activation quantization~\cite{hwgq,binaryduo,dorefa}.
In addition, when quantizing the activation, the accuracy loss is marginal until 2-bit quantization, but a significant accuracy drop occurs when quantizing the activation to 1-bit precision~\cite{binaryduo,abc-net,wrpn}.
Previous works tried to explain the sharp accuracy drop from 2-bit activation to 1-bit case based on the gradient mismatch problem caused by the non-differentiable binary activation function~\cite{bnn+,bireal}.
Since the quantization functions are non-differentiable, gradients cannot propagate through the quantization layer in the back-propagation process.
Therefore, previous works used straight-through-estimator (STE) to compute the approximate gradient on non-differentiable layers~\cite{ste,bnn}.
While STE enables back-propagation through quantization layers, the discrepancy between the actual function and the approximated function causes gradient mismatch problem.
Especially in BNNs, sign function is used to binarize activations and is usually approximated as hardtanh function in back-propagation.
Compared to other multi-bit quantization functions, sign function shows more severe gradient mismatch which leads to sharp accuracy degradation.
Hence, several works tried to design better approximation function~\cite{bnn+,bireal} or to reduce the gradient mismatch using neuron coupling~\cite{binaryduo}.

In this work, we argue that there is another reason for the poor performance of BNN in addition to the gradient mismatch.
We speculate that the symmetry of the sign function is also partly responsible for the degradation of BNN performance.
Most DNN models use ReLU activation function instead of sigmoid or Tanh functions.
While the output distributions of sigmoid and Tanh functions are symmetric with respect to zero, ReLU function replaces all the negative values to zero so that the distribution of ReLU output is highly skewed. When quantizing activations to 2-bit or a higher precision, ReLU-based quantization functions are usually used~\cite{pactsawb,pact,lsq,dorefa}.
In other words, multi-bit quantization functions also output unbalanced activation distributions similar to ReLU function.
However, when binarizing activations, sign function is used, and hence the distribution of the binary activation becomes symmetric (or balanced).
We show that a model with unbalanced activation distribution performs better than that with balanced activation distribution.
We first show that the claim is valid even in the full precision activation case by comparing hardtanh and ReLU6 activation functions.
We then show that the performance of BNN can be improved by simply shifting the threshold of the sign function.
We also analyze the effect of training the threshold of the sign function and show that the thresholds cannot be trained efficiently through back-propagation.
Our contributions can be summarized as follows:
\begin{itemize}
    \item To the best of our knowledge, we are the first to report that the unbalanced distribution of binary activation actually helps improve the accuracy of BNNs.
    \item We propose to shift the thresholds of binary activation functions to make the distribution of binary activation unbalanced. 
    \item Experimental results show that the accuracy of previous BNN models can be improved at almost no cost by simply shifting the threshold of the binary activation function.
\end{itemize}

\section{Related Work}
\subsection{Network binarization}
There have been several approaches to quantize weights and/or activations into 1-bit precision.
Courbariaux et al.~\cite{binaryconnect} demonstrated binary weight networks which can be successfully trained on small datasets such as CIFAR-10 and SVHN.
Hubara et al.~\cite{bnn} further proposed BNN in which both weights and activations are binarized.
To apply back-propagation through a sign function which is non-differentiable, the straight-through-estimator (STE) concept was used~\cite{ste}.
Rastegari et al.~\cite{xnor} proposed XNOR-Net which uses real-valued scaling factors when binarizing weights and activations and demonstrated acceptable accuracy on large scale ImageNet dataset.
When binarizing ResNet models, real-valued shortcut connections play a critical role in propagating high resolution information.
Liu et al.~\cite{bireal} proposed to use additional shortcut connections so that a shortcut connection exists for every binary convolution layer.

\subsection{Training quantization parameters}
Recent studies suggested to train the quantization intervals and ranges using back-propagation to improve the accuracy of quantized neural networks.
Choi et al.~\cite{pactsawb,pact} proposed the parameterized clipping activation function (PACT) in which the clipping range is trained using back-propagation.
While only the clipping range was trained in PACT, several following works proposed to train both quantization interval and range~\cite{lsq,qil,lqnet}.
Note that these works focused on training multi-bit networks and hence did not report results on BNNs.
For BNNs, several recent works proposed to train the threshold of binary activation function.
Liu et al.~\cite{reactnet} proposed to use trainable activation functions so that the distribution of the activation can be balanced.
Wang et al.~\cite{sibnn} proposed the trainable binarization which learns the threshold as well as the gradient clipping range used in back-propagation.
We, however, observed results contrary to the findings from previous works on trainable threshold.
Our experimental results show that the bias term in Batch Normalization (BN) layer is already learning the threshold and therefore the effect of training the threshold of binary activation is limited. Details will be described in Sec.~\ref{sec:learnable}.

\subsection{Managing activation distribution}
Since there are only two values available for activation in BNNs, the distribution of binary activation plays a critical role in BNNs.
There have been few works that tried to manipulate the distribution of binary activation to improve the accuracy of BNNs.
Ding et al.~\cite{distloss} proposed to regularize the distribution of pre-activation values to tackle the difficulties that occurred during training BNNs.
The work mostly focused on avoiding extreme cases such as the case when all the pre-activation values have the same sign.
Liu et al.~\cite{reactnet} proposed to reshape the distribution of binary activation using trainable thresholds.
Using the trainable activation functions, they made the distribution of binary activation more balanced.

\subsection{Additional activation function}
Another simple yet effective technique to improve the accuracy of BNNs is to use an additional activation function (e.g. PReLU) between the binary convolution layer and the following BN layer.
Rastegari et al.~\cite{xnor} mentioned that inserting a ReLU function after the binary convolution helps training BNNs.
Tang et al.~\cite{aaai} proposed to use PReLU instead of ReLU to absorb the weight scale factors. 
Bulat et al.~\cite{bulat2017,bulat2019} explained that increasing the nonlinearity in BNNs helps training, and compared the effects of ReLU and PReLU. 
Based on these findings, many recent works utilized additional activation layers in their models~\cite{reactnet,rtb,leastsquares}.

\section{Method}
In this section, we introduce how to improve the accuracy of BNNs using unbalanced activation distribution.
We first discuss the activation functions in conventional full precision models to describe the motivation for using unbalanced activation distribution (Sec.~\ref{sec:motivation}).
Then we show how to make the distribution of the binary activation unbalanced using threshold shifting in BNNs and how much the proposed technique can improve the accuracy on various benchmarks (Sec.~\ref{sec:biasing}).
Experimental results on ImageNet dataset will be given to show that the accuracy of various existing BNN models can be improved by simply shifting the threshold of binary activation functions (Sec.~\ref{sec:imagenet}).
After that, we discuss the ineffectiveness of the methodology to train the threshold of binary activation in BNNs (Sec.~\ref{sec:learnable}).
We also show that the additional activation functions (i.e. ReLU or PReLU) used in recent BNN models make the activation distribution unbalanced thereby helping to improve the accuracy (Sec.~\ref{sec:relu}).

\subsection{Accuracy gap between hardtanh and ReLU6}
\label{sec:motivation}
\begin{figure}
    \centering
    \includegraphics[width=\linewidth]{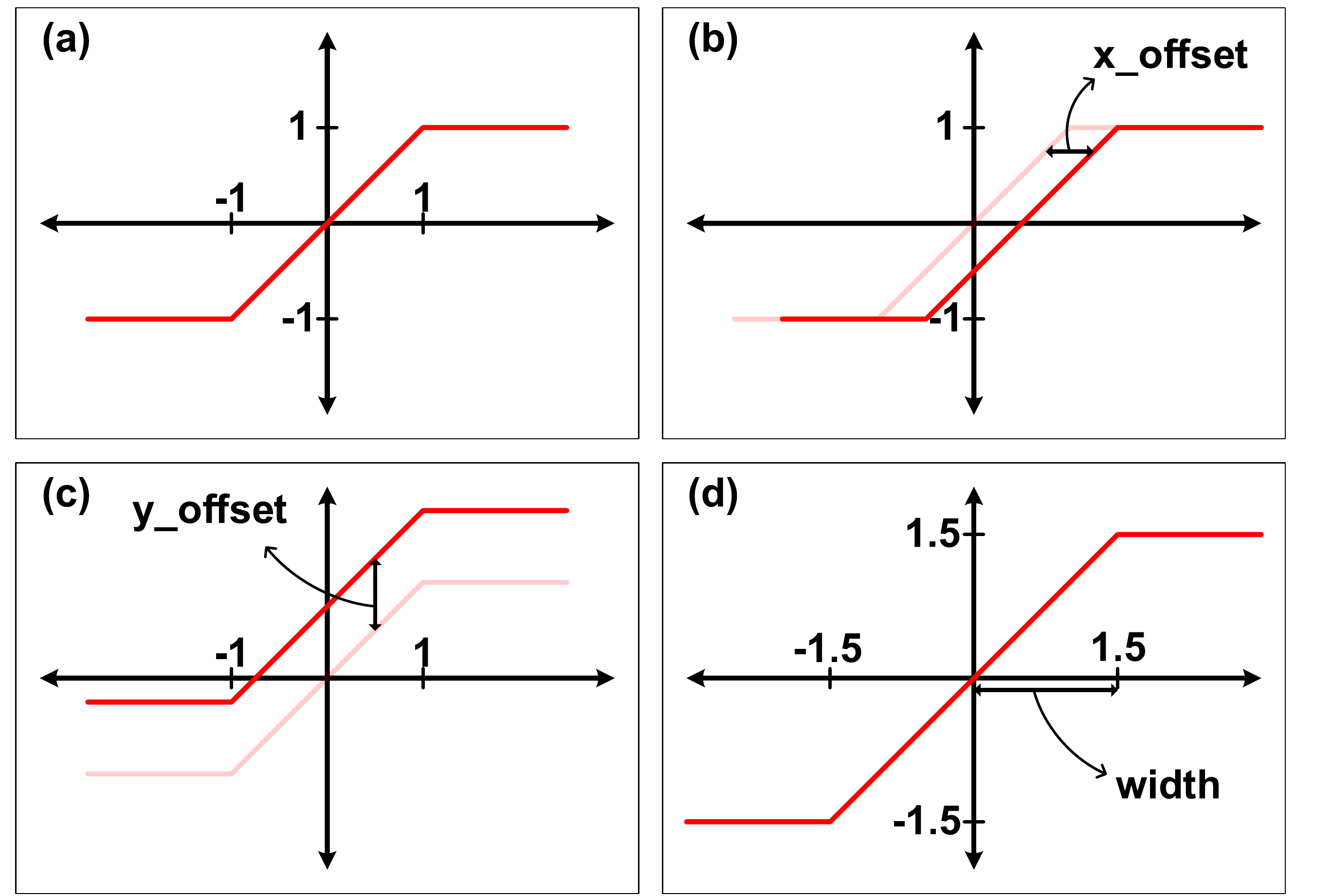}
    \caption{(a) Original hardtanh function. Hardtanh function shifted along (b) the x-axis by x\_offset and (c) the y-axis by y\_offset. (d) Hardtanh function with the increased range.}
    \label{fig:hardtanh}
\end{figure}

Before ReLU activation function was proposed, sigmoid or Tanh functions had been used as an activation function.
It is widely known that ReLU works better because it solves the gradient vanishing problem that occurs in sigmoid or Tanh functions~\cite{relu}.
However, the gradient vanishing problem is largely diminished by using BN layers in recent models~\cite{bn}, so there might be other reasons for the success of the ReLU function.
We suspect that another main reason for the good performance of ReLU is that the output distribution of the ReLU function is highly skewed.
While the output distributions of sigmoid and Tanh functions are symmetric with respect to each mean value, ReLU makes all the negative inputs to zero and passes the positive inputs.
Therefore, the output distribution of the ReLU function is positively skewed (mean value is larger than the median value).
We conducted a few experiments to monitor whether the unbalanced activation distribution due to the nature of ReLU function helps to improve accuracy.
To make the problem simple, we compare the performance of hard hyper-tangent (hardtanh) and ReLU6 functions instead of ReLU.
The ReLU6 is a slight variant of the ReLU function where its positive outputs are clipped to 6. 
The hardtanh function is shown in Fig.~\ref{fig:hardtanh}a.
It has been reported that the performance of ReLU6 is as good as that of ReLU or sometimes even better~\cite{relu6}.
Also, ReLU6 and hardtanh have very similar shape and ReLU6 can be thought as a shifted-and-scaled form of hardtanh.
As shown in Fig.~\ref{fig:hardtanh}, we modified the hardtanh function in three different ways; (b) shifting along the x-axis, (c) shifting along the y-axis and (d) increasing the range.
Note that the hardtanh function is identical to the ReLU6 function when it is shifted along the x- and y-axis by 3 and its range is increased by 3 times.
We trained vgg-small model~\cite{vgg} on CIFAR-10 dataset with different activation functions.
The vgg-small model has 4 convolution layers with 64, 64, 128, 128 output channels in sequence followed by 3 fully-connected layers with 512 neurons.
We trained the model in the same condition with 10 different seeds and report the mean and the standard deviation of the test accuracy.
Detailed setup for the training is described in the supplementary material.
When ReLU6 function is used as the activation function, 89.21\% of test accuracy was achieved while 88.55\% was achieved when hardtanh function is used without any modification.
When the hardtanh activation function is modified as described in Fig.~\ref{fig:hardtanh}, the test accuracy of the model also changes.
\begin{figure}
    \centering
    \includegraphics[width=\linewidth]{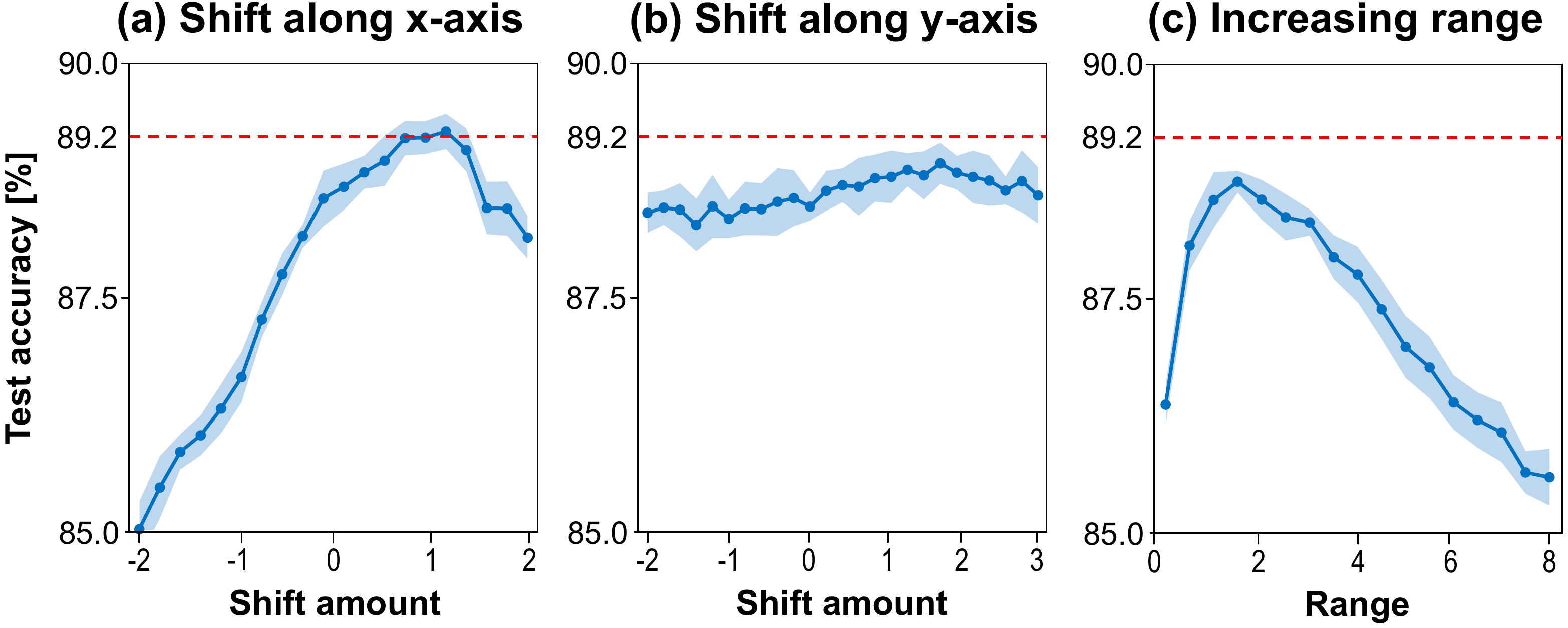}
    \caption{Test accuracy of vgg-small model with modified hardtanh activation functions. Red dotted line represents the test accuracy of the model with ReLU6 activation function. The mean (line) and standard deviation (shade) of 10 runs are plotted.
    }
    \label{fig:vggsmall_hardtanh}
\end{figure}
\begin{figure}
    \centering
    \includegraphics[width=\linewidth]{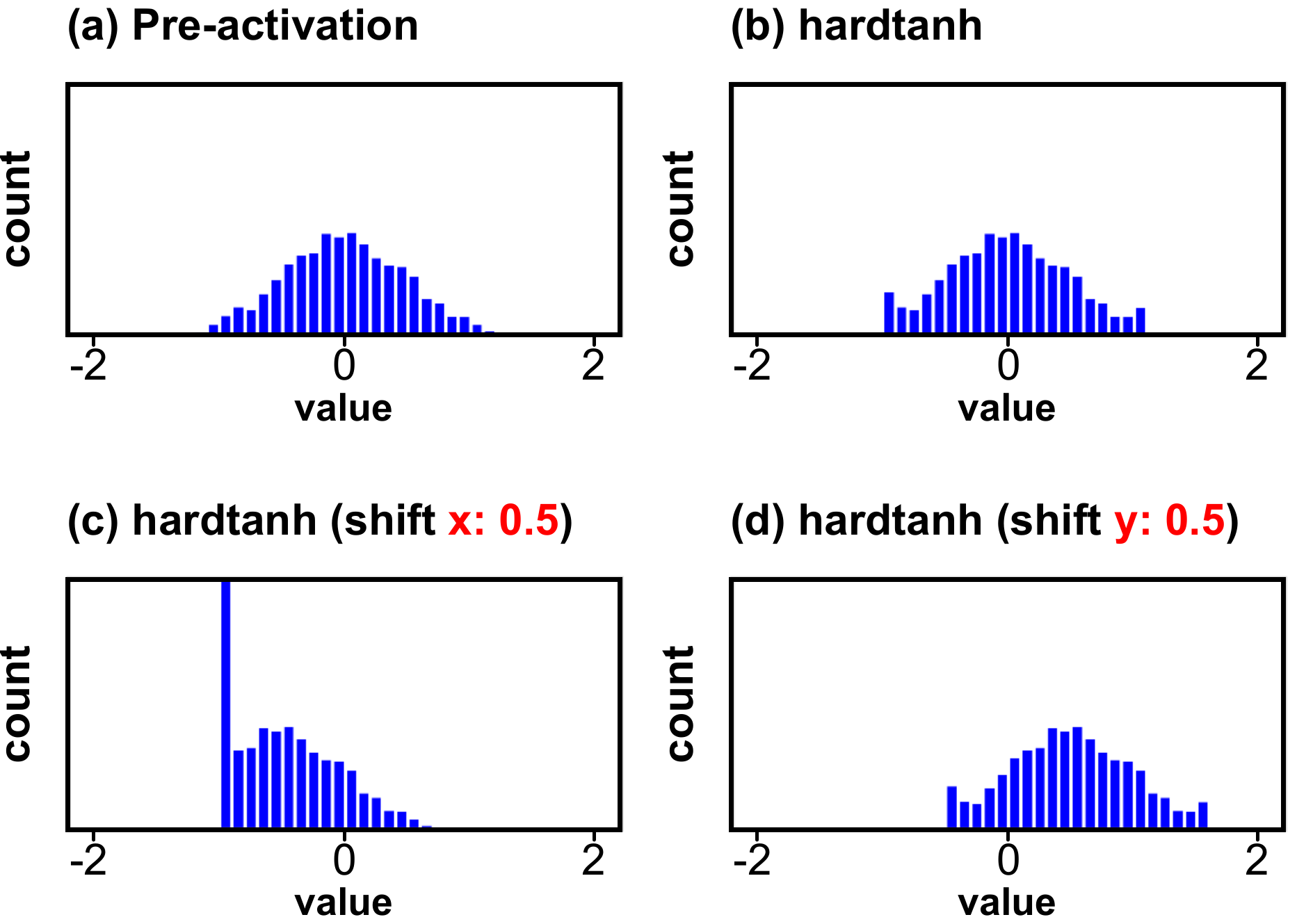}
    \caption{Distributions of (a) pre-activation, (b) output of hardtanh activation function, (c) output of hardtanh activation function shifted to the right along the x-axis by 0.5, and (d) output of hardtanh activation function shifted up along the y-axis by 0.5. }
    \label{fig:skew_dist}
    \vspace{-2mm}
\end{figure}
Fig.~\ref{fig:vggsmall_hardtanh} shows the change in the test accuracy depending on the amount of shift along (a) the x-axis or (b) y-axis or (c) increase in the range of the hardtanh activation function.
We observe that shifting the hardtanh activation function along the x-axis increases the accuracy substantially (Fig.~\ref{fig:vggsmall_hardtanh}a).
The highest test accuracy was obtained when shifting the hardtanh function in a positive direction along the x-axis by 1.2.
On the other hand, shifting the hardtanh activation along the y-axis or increasing the range does not improve the accuracy as much as shifting the hardtanh along the x-axis (Fig.~\ref{fig:vggsmall_hardtanh}(b and c)).
When the hardtanh function is shifted to the right along the x-axis, the number of negative outputs increases and the output distribution becomes positively skewed (Fig.~\ref{fig:skew_dist}c).
Note that shifting the hardtanh function along the y-axis also makes the output activation distribution to have a non-zero mean.
However, it only shifts the activation distribution and the distribution is not skewed in such a case (Fig.~\ref{fig:skew_dist}d).
Therefore, we think that the higher performance of ReLU activation function compared to the hardtanh partly comes from the unbalanced distribution of activation outputs.
Interestingly, the effect of the activation distribution becomes even more noticeable in BNNs when the hardtanh function is replaced by the sign function.

\subsection{BNNs with unbalanced activation distribution}
\label{sec:biasing}
As described in the previous section, breaking the balance of the activation distribution by shifting the activation function along the x-axis helps improve the accuracy of a model.
When quantizing activations, previous multi-bit quantization methods~\cite{hwgq,pactsawb,pact,lsq,syq,qil,wrpn,dorefa} used ReLU-based quantization which outputs unbalanced activation distributions.
However, in BNNs, previous works used sign function for binary activation function thus the distribution of binary activation is balanced~\cite{xnor++,bnn+,distloss,bnn,bireal,xnor}.
Note that in BNNs, there are only two output values (+1 or -1) and the balanced distribution means that the ratio of +1 to -1 is close to 1:1.
We believe that the poor performance of BNNs observed in previous literature~\cite{abc-net,wrpn} is partly due to the use of the sign function.
Since shifting the hardtanh function along the x-axis improved the test accuracy in the full precision model, we also tried shifting the sign function along the x-axis.
Note that shifting the sign function is the same as just changing the threshold value of the activation function.
\begin{figure*}
    \centering
    \includegraphics[width=\linewidth]{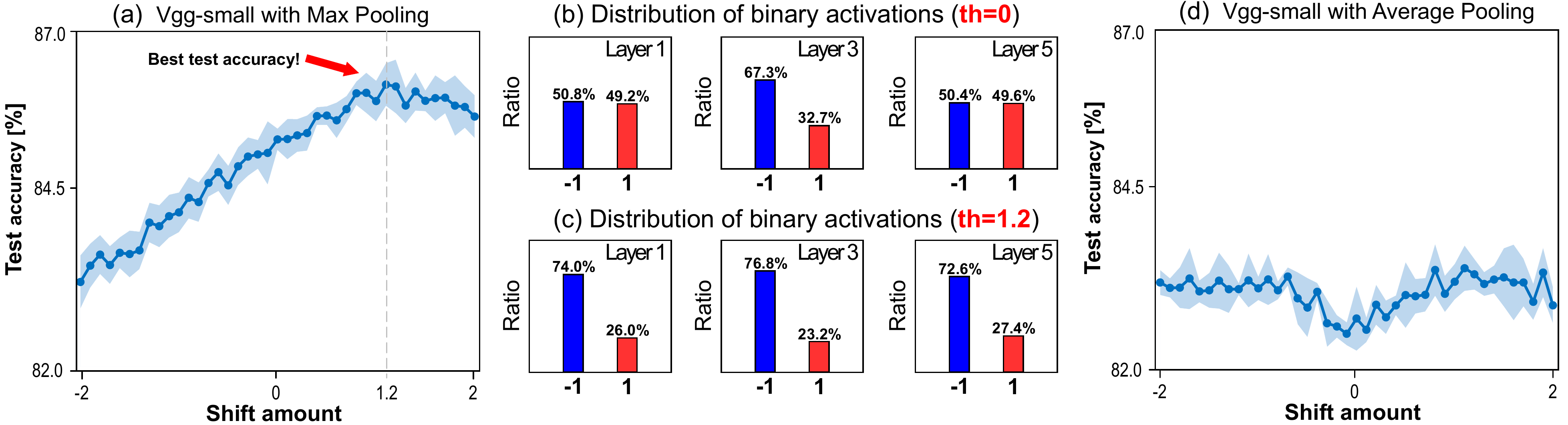}
    \caption{(a) Test accuracy vs. threshold shift for vgg-small model. (b) Distribution of the binary activation of the first, third, and fifth activation layer when threshold shift = 0. (c) Distribution of the same data with threshold shift = 1.2. (d) Test accuracy vs. threshold shift when Max Pooling layers are replaced by Average Pooling layers.}
    \label{fig:vggsmall_sign}
\end{figure*}
As expected, shifting the activation function along the x-axis increases the accuracy of BNNs also (Fig.~\ref{fig:vggsmall_sign}a).
To check the effect of the shape of the distribution of activation outputs, we observed the distributions of the binary activation in different layers. %data before and after the binary activation layer.
Fig.~\ref{fig:vggsmall_sign}b shows the distribution of the binary activations of the first, third, and fifth activation layer when the original sign function is used.
For the first and the fifth activation layer, the distribution of pre-activation is Gaussian-like with zero mean and hence the ratio of +1 to -1 in binary activation is close to 1:1.
In the case of the third activation layer, the distribution of pre-activation values does not have zero mean due to the preceding Max Pooling layer even when the original sign function is used.
Note that the convolution results go through the Max Pooling layer first and then a BN layer before the binary activation layer.
The Max Pooling layer makes the distribution positively skewed, thus the mean of the distribution becomes larger than the median of the distribution.
Since the BN layer centers the distribution based on the mean value, there exist more negative values than positive values in pre-activation values.
As a result, the number of -1 is larger than that of +1 in the binary activation after the third activation layer (Fig.~\ref{fig:vggsmall_sign}b).
However, the output distributions of all other layers which do not come after Max Pooling layer are balanced.
When the threshold of binary activation function is shifted by 1.2, which results in the best accuracy, the distribution of binary activation changes.
Fig.~\ref{fig:vggsmall_sign}c shows the distribution of the binary activations when the threshold is shifted by 1.2.
As expected, the output distributions of all the binary activation layers become unbalanced with the shifted threshold.
Note that even though the distribution of binary activation in a layer is unbalanced, the imbalance does not propagate through layers because the binary weights in the following binary convolution layer are zero-centered.

\subsubsection{Effect of Max Pooling}
Another interesting observation from the results of shifting the activation function along the x-axis was that the accuracy improved only when the activation function was shifted in a positive direction.
Since the Max Pooling layer is the only layer that gives asymmetry in the model, we suspect that Max Pooling causes the accuracy difference between shifting of the threshold to the positive direction and the negative direction.
To gauge the effect of the Max Pooling, we replaced the Max Pooling layers with Average Pooling layers which do not have asymmetric effects.
Fig.~\ref{fig:vggsmall_sign}d shows the training results.
Since the model with Average Pooling is symmetric, the effect of shifting the activation function is also symmetric and we can observe that the accuracy is improved when the threshold is shifted away from zero in either direction.

\subsubsection{Effect of other conditions}
In this section, we investigate the effect of threshold shifting with various (1) datasets, (2) models, (3) initialization methods, and (4) optimizers to show that the proposed technique is not limited to the specific condition or benchmark.

\begin{figure}
    \centering
    \includegraphics[width=\linewidth]{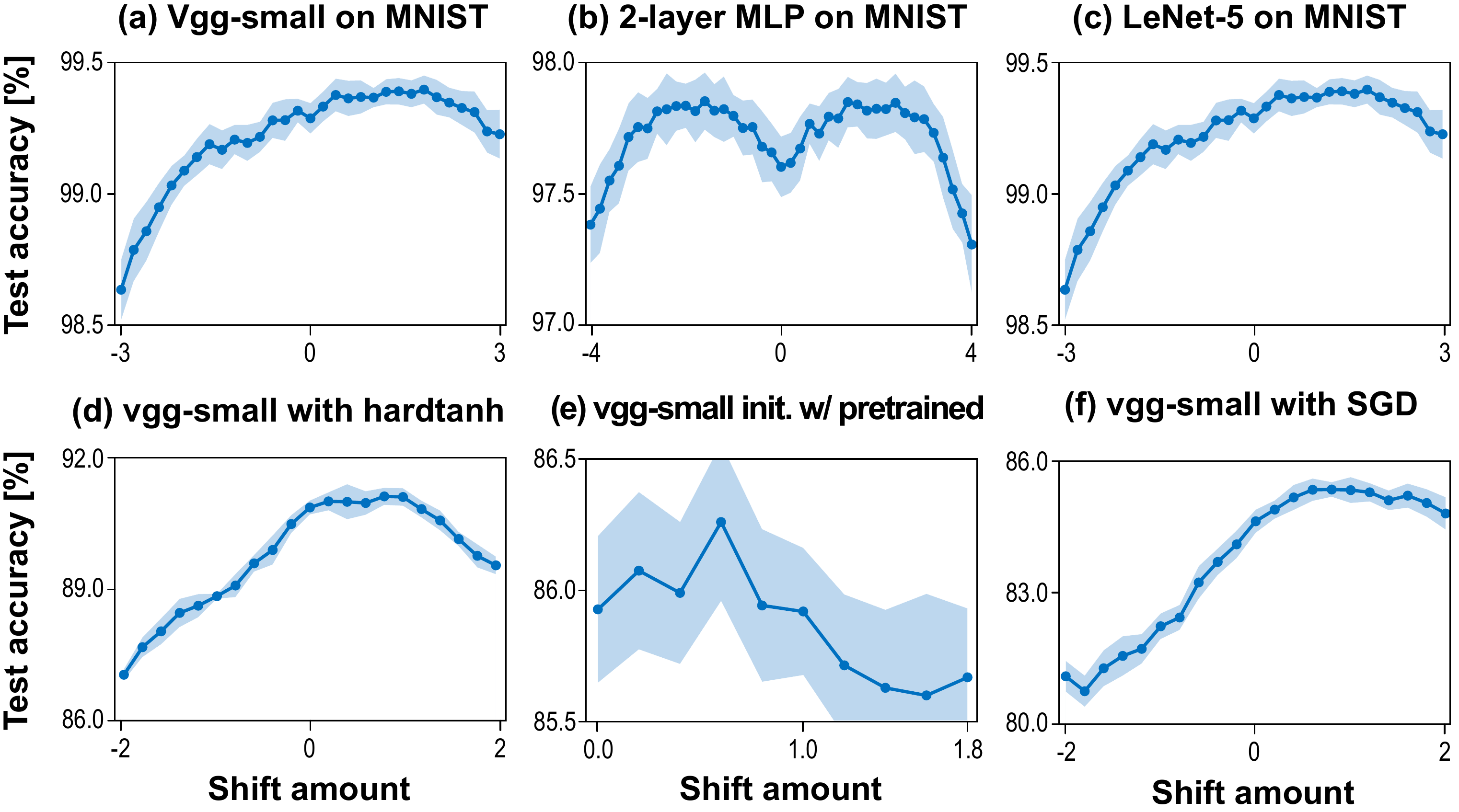}
    \caption{Test accuracy vs. threshold shift for (a) binary vgg-small model on MNIST dataset, (b) binary 2-layer MLP model on MNIST dataset, (c) binary LeNet-5 on MNIST dataset, (d) full-precision vgg-small model on CIFAR-10 dataset, (e) binary vgg-small model initialized with full-precision pretrained model, and (f) binary vgg-small model optimized using SGD.}
    \label{fig:others}
\end{figure}

\noindent\textbf{Dataset.} We first changed the dataset to MNIST.
To solely investigate the dependency on dataset, we used the same vgg-small model which we used for CIFAR-10 in Sec.~\ref{sec:biasing}.
For MNIST dataset, we trained each model with 30 different random seeds and average results are presented.
As shown in Fig.~\ref{fig:others}a, the accuracy of the vgg-small model can also be improved by shifting the threshold when trained on MNIST dataset.
We also evaluate the ImageNet dataset and the results will be described in the Sec.~\ref{sec:imagenet_result}.

\noindent\textbf{Model architecture.} 
We also evaluated two additional model architectures (2-layer MLP and LeNet-5) to verify the effect of threshold shifting.
Fig.~\ref{fig:others}b and c show the result of MLP and LeNet-5, respectively.
Although the accuracy improvement with threshold shifting is relatively small ($\sim$0.2\%), a clear trend is observed in the test accuracy as the activation functions are shifted along the x-axis in both models.
Note that the trend in the MLP model is symmetric because the model do not have the Max Pooling layer in it.
We also conducted experiments on much larger and complex models (AlexNet and ResNet), and the results will be described in the Sec.~\ref{sec:imagenet_result}.

\noindent\textbf{Initialization method.}
In all the previous experiments, we trained the models from scratch using Xavier normal initialization~\cite{xavier}.
In recent literature~\cite{bireal,rtb}, pretrained full-precision models are often used to initialize BNN models.
We also demonstrate that the proposed technique is effective with such an initialization method.
As shown in Sec.~\ref{sec:motivation}, the accuracy of a model with hardtanh function can also be improved by shifting the activation function to the right along the x-axis.
Since pretrained full-precision models typically use hardtanh function instead of ReLU function to minimize the mismatch with the sign activation function in the initialization stage, we shifted the hardtanh function in the full-precision model and used the results as the initialization points.
Fig.~\ref{fig:others}d shows the training results of vgg-small model with full-precision weight and activation.
We initialized the BNN version of vgg-small model with the pretrained model in which the hardtanh function is shifted by 1.
While using the same shift amount for binary activation and hardtanh function seems natural, there might be a better shift amount for the binary activation other than the shift amount for the pretrained model.
Therefore, we searched for the optimal shift amounts for the pretrained model and BNN model separately.
Fig.~\ref{fig:others}e shows the trend of test accuracy versus the threshold shifting amount.
While the pretrained model used 1.0 as the shift amount, the best shift amount for the BNN model was 0.6 so using different amounts of shift for the pretrained model and the BNN produces higher accuracy.

\noindent\textbf{Optimizer.}
Although most previous BNN training methods used Adam optimizer~\cite{adam}, we also tested SGD with momentum to see the sensitivity to the optimizer.
Fig.~\ref{fig:others}f shows that a similar trend is maintained with SGD while the accuracy is slightly degraded compared to the Adam case.

\subsection{Experiments on ImageNet}
\label{sec:imagenet}
In this section, we apply the threshold shifting technique to various previous BNN models on ImageNet dataset and show that the technique can be added to existing BNN training techniques without requiring any modification other than the simple threshold shift.
We also introduce a simple method to find the appropriate shifting amount for each model to avoid time-consuming search for optimal threshold shift for large neural networks.

\subsubsection{Finding shift amount}
\begin{figure}
    \centering
    \includegraphics[width=0.85\linewidth]{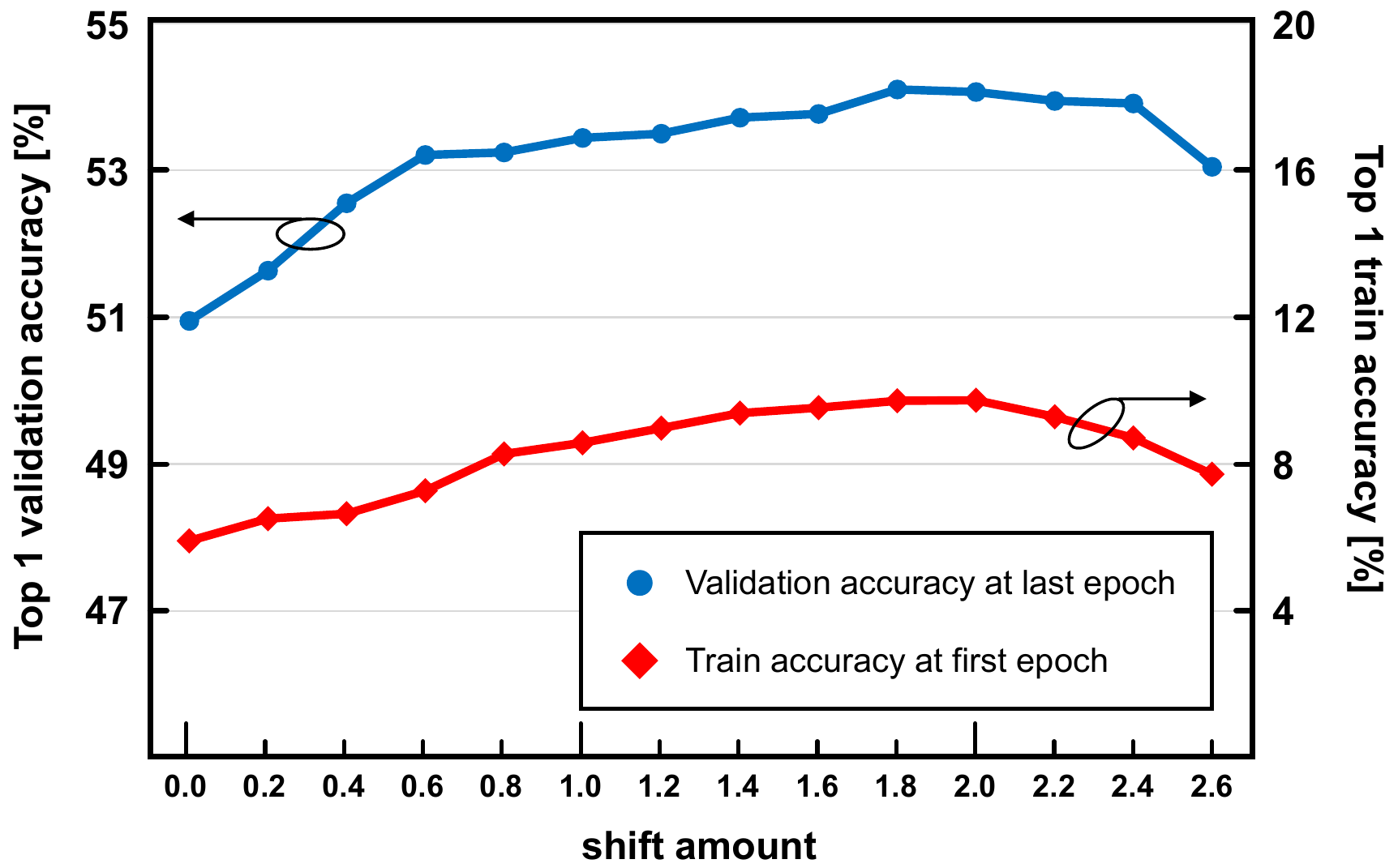}
    \caption{Top-1 train accuracy at the first epoch and Top-1 validation accuracy at the end of training with various threshold shift amount. The two graphs show similar trends.}
    \label{fig:train_vs_valid}
    \vspace{-2mm}
\end{figure}
As shown in previous results, the best shift amount differs across models and datasets.
To find the optimal shift amount, we propose to use the training accuracy in earlier training steps.
For example, we choose the best shift amount based on the train accuracy at the end of the first epoch for ImageNet dataset.
This approach is valid because the trend in the train accuracy at the first epoch and that in the validation accuracy after training is finished are similar enough.
Fig.~\ref{fig:train_vs_valid} shows the two trends when training XNOR-Net (ResNet-18)~\cite{xnor} on ImageNet dataset.
The shapes of the two trends resemble each other, and both trends show the peak performance when their shift amounts are around 2.0.
Using the proposed method, we can find the optimal value for the threshold shift easily without training the model to the end for every shift amount.

\subsubsection{Results on ImageNet}
\label{sec:imagenet_result}
Using the proposed method to find optimal threshold shift, we trained several previous BNN models for ImageNet dataset with the threshold shifting technique.
\begin{table*}[t]
    \centering
    \caption{Accuracy of previous BNN models trained on ImageNet dataset with and without the threshold shifting technique. $^\dagger$For Bi-Real-Net, the shift amounts for the pretrained model and the BNN model are shown together.}
    \begin{tabular}{c|cc|c|cc|cc}
         \toprule
         \multirow{2}{*}{Model} & \multicolumn{2}{c|}{Baseline accuracy [\%]} & \multirow{2}{*}{Shift amount} & \multicolumn{2}{c|}{Accuracy w/ shift [\%]} & \multicolumn{2}{c}{Accuracy gain [\%]} \\
                                    &\;\;\;\;Top-1  & Top-5\;\;   &           & \;\;\;\;Top-1   & Top-5\;\; & \;\;\;Top-1 & Top-5\;\;\\
         \hline
         BNN (AlexNet)              &\;\;\;\;41.5   & 66.1\;\;    & 0.6       & \;\;\;\;42.1    & 66.6\;\;  & \;\;\;0.6   & 0.5\;\;\\
         XNOR-Net (AlexNet)         &\;\;\;\;44.4   & 68.6\;\;    & 0.8       & \;\;\;\;45.6    & 69.6\;\;  & \;\;\;1.2   & 1.0\;\;\\
         XNOR-Net (ResNet-18)       &\;\;\;\;51.2   & 74.9\;\;    & 2.0       & \;\;\;\;54.2    & 77.6\;\;  & \;\;\;3.0   & 2.7\;\;\\
         Bi-Real-Net (ResNet-18)    &\;\;\;\;56.1   & 79.1\;\;    & (2.0, 1.6)$^\dagger$   & \;\;\;\;57.2    & 80.2\;\;  & \;\;\;1.1   & 1.1\;\;\\
         Bi-Real-Net (ResNet-34)    &\;\;\;\;61.9   & 83.9\;\;    & (2.0, 1.8)$^\dagger$   & \;\;\;\;62.8    & 84.5\;\;  & \;\;\;0.9   & 0.6\;\;\\
         \bottomrule
    \end{tabular}
    \label{tab:imagenet}
\end{table*}
Table~\ref{tab:imagenet} shows how much accuracy improvement can be achieved by the proposed threshold shifting technique on different BNN models.
We first reproduced the baseline accuracy of previous works with less than 0.3\% accuracy difference.
With the same hyper-parameter setting, we only changed the threshold of binary activation functions in each model.
The shift amount for each model is shown in the Table~\ref{tab:imagenet}.
The results indicate that the proposed threshold shifting technique also works well for large models on ImageNet dataset.
In addition, the proposed threshold shifting technique can be easily combined with other BNN training methods (i.e. weight scaling factor and double skip connection).

\subsection{Effect of training threshold}
\label{sec:learnable}
In several recent works, it has been proposed to train the interval and range of quantization functions via back-propagation~\cite{pact,lsq,qil,lqnet}.
Training the threshold of binary activation function in BNNs has been also proposed recently~\cite{reactnet,sibnn}.
The threshold training approaches are based on the belief that the best threshold value for each binary activation function can be found using back-propagation.
However, here we show that training the threshold has a limited effect on the performance of BNNs.

\subsubsection{Batch normalization bias}
The most important reason why the threshold training is not effective is that the role of trainable threshold is already covered by the bias values in the BN layer which comes right before the binary activation layer.
The compute process of a BN layer and the following binary activation layer can be represented as 
\begin{equation}
    \label{eq:bn-sign}
    \begin{split}
    Y=
    \begin{cases}
    -1 & \quad \text{if}\;\; \gamma\frac{X-\mu}{\sigma}+\beta \leq th\\
    +1 & \quad \text{if}\;\; \gamma\frac{X-\mu}{\sigma}+\beta > th
    \end{cases}.
    \end{split}
\end{equation}
Here, $X$ and $Y$ are inputs to the BN layer and output from the binary activation layer, respectively.
$\mu, \sigma, \gamma,$ and $\beta$ are mean, standard deviation, weight, and bias of the BN layer, and $th$ is the trainable threshold.
As shown in the Eq.~\ref{eq:bn-sign}, the bias term in BN serves exactly the same role as the trainable threshold.
Both of them are initialized to zero in the beginning and their gradients are values of the same magnitude and the opposite sign.
\begin{figure}
    \centering
    \includegraphics[width=0.85\linewidth]{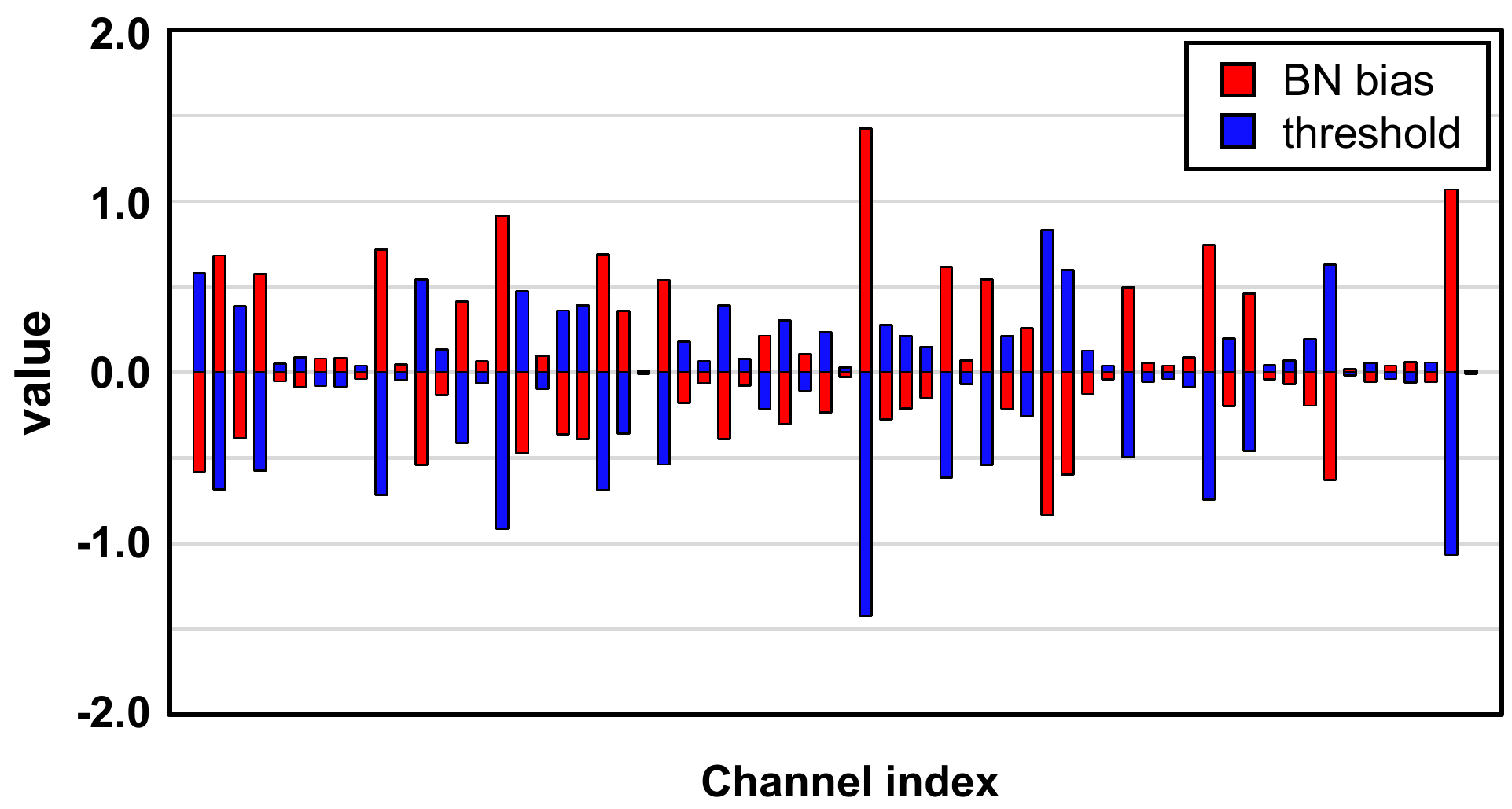}
    \caption{Comparison of the BN bias values in the second BN layer and the threshold values in the following activation layer of the vgg-small model. The thresholds are trained following the approach suggested in \cite{reactnet}.}
    \label{fig:bias_vs_threshold}
    \vspace{-2mm}
\end{figure}
We trained the vgg-small model with trainable thresholds following \cite{reactnet} and observed the bias values of the second BN layer and the threshold values of the following binary activation layer.
Fig.~\ref{fig:bias_vs_threshold} shows the values of BN biases and the trained thresholds in the vgg-small model.
For each channel, the BN bias and the threshold have the same values with an opposite sign.
Therefore, training the thresholds of binary activation functions means nothing more than doubling the learning rate of BN biases.
Please refer to section S2 in the supplementary material for more discussion.
\subsubsection{Limited learning capability}
\begin{figure}
    \centering
    \includegraphics[width=0.95\linewidth]{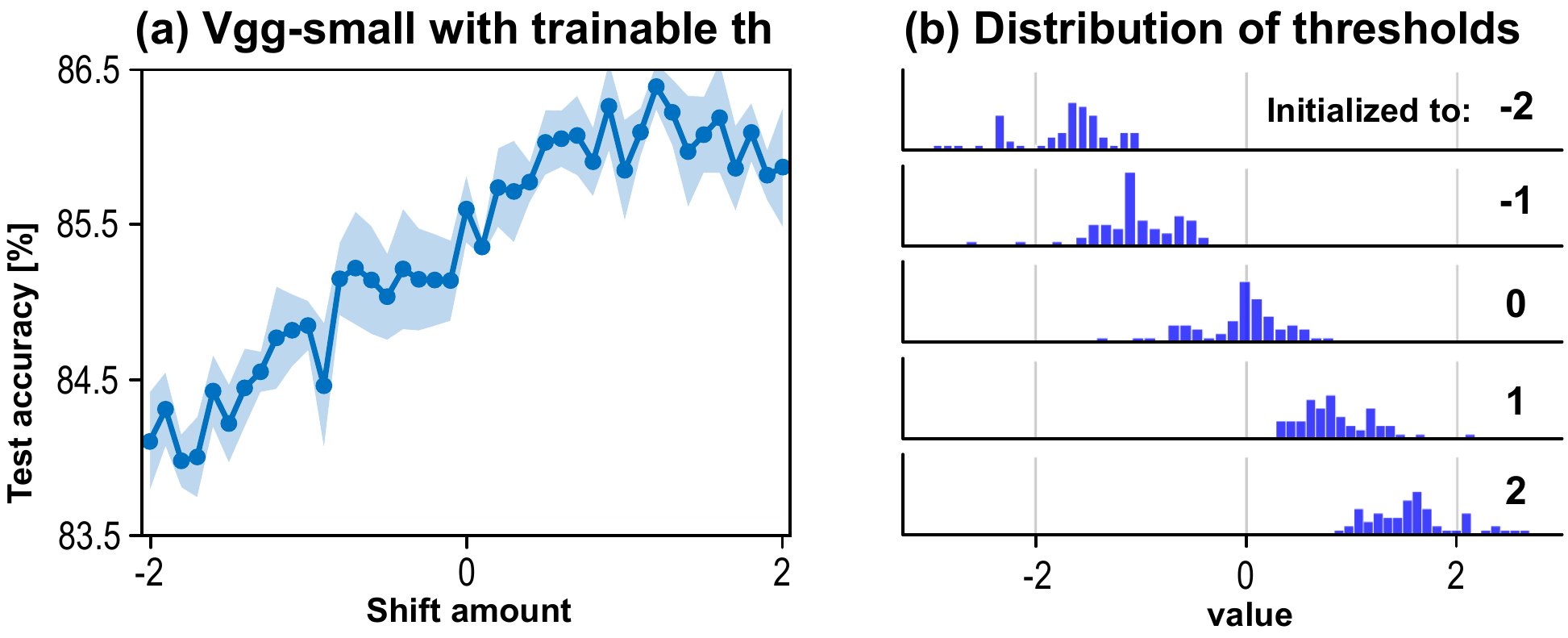}
    \caption{(a) Test accuracy of vgg-small model with trainable thresholds with various initial threshold values. (b) Distribution of effective thresholds ($th-\beta$) of the second binary activation layer after training is finished. Each distribution from top to bottom represents the case when the thresholds are initialized to -2, -1, 0, 1, and 2, respectively.}
    \label{fig:learnable}
    \vspace{-3mm}
\end{figure}
While the shifted threshold values in the proposed method can also be absorbed by the bias values in BN after a training is finished, it differs from the trainable threshold method in that the fixed shift amount of the binary activation function serves as an initialization point as well.
If the BN bias and the threshold can be trained to an optimal point via back-propagation from an arbitrary initial point, the threshold shifting technique might not be as effective.
We, however, found that the final BN bias and the trainable threshold values heavily depend on the initial values, and hence initializing them with the proposed threshold shifting technique strongly affects the training performance.
Fig.~\ref{fig:learnable}a shows the training result of vgg-small model with trainable threshold. 
To see the effect of initialization values, we varied the amount of the threshold shift and applied the trainable threshold methods.
Similar to the result with a fixed threshold, the test accuracy strongly depends on the initialization point (or threshold shifting amount).
We further analyzed the distribution of effective threshold ($th-\beta$) after training is finished for five different initialization values.
As shown in Fig.~\ref{fig:learnable}b, the final distribution of the effective threshold strongly depends on the initial values and do not change much from the initial points.
For example, the first case (top) in Fig.~\ref{fig:learnable}b is when the threshold of the binary activation function is set to -2 and the BN bias is initialized to 0.
Even though the thresholds of the binary activation functions, which are initially -2, are trainable, back-propagation does not train them to the optimal value which is close to 1.

\subsection{Effect of additional activation function}
\label{sec:relu}
Recently, several works proposed to use additional activation function (i.e. PReLU) after convolution layer in BNNs~\cite{bulat2017,bulat2019,reactnet,rtb,leastsquares,aaai}.
While the performance improvement by the additional activation function was significant, the reason for the accuracy improvement is not clearly understood yet.
Previous works have described that BNNs usually lack nonlinearity in the model due to the simple activation functions and the additional activation functions give more nonlinearity to the model~\cite{bulat2017,xnor}.
In addition to the increased nonlinearity, we observed that the accuracy improvement by additional activation function is also related to the unbalanced activation distribution.
\begin{figure}
    \centering
    \includegraphics[width=0.95\linewidth]{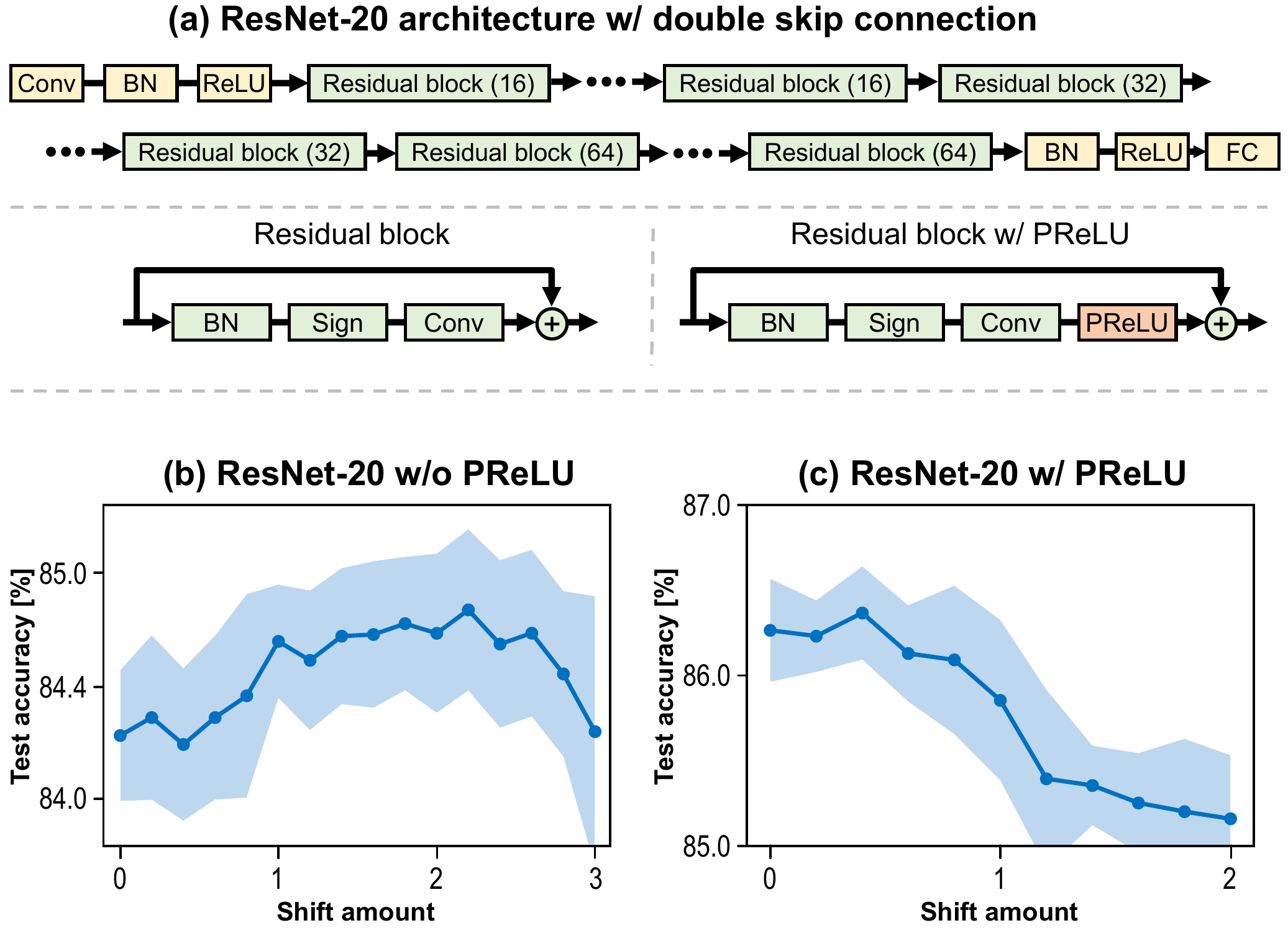}
    \caption{(a) Model description of ResNet-20 with double skip connection. ResNet-20 with PReLU model has an additional PReLU layer at the end of every residual block. Test accuracy of ResNet-20 (b) without and (c) with an additional PReLU layer after every binary convolution layer. 
    }
    \label{fig:prelu}
    \vspace{-5mm}
\end{figure}
We first evaluate the effect of additional PReLU layers after binary convolution layers in the ResNet-20 model.
For experiments, we used ResNet-20 model with double skip connections~\cite{bireal} as shown in Fig.~\ref{fig:prelu}a.
The model consists of the first convolution layer, 18 residual blocks, and the last fully-connected layer.
In the ResNet-20 model with additional PReLU layers, a PReLU layer is inserted at the end of every residual block as shown in Fig.~\ref{fig:prelu}a.
The effect of threshold shifting on ResNet-20 model with and without the additional PReLU layers is shown in Fig.~\ref{fig:prelu}b and c, respectively.
While the accuracy is improved by shifting the threshold of binary activation function in the ResNet-20 model, the threshold shifting degrades the accuracy when additional PReLU layers are used jointly.
The additional PReLU layers already play a role in distorting the activation distribution, and hence further shifting the threshold of binary activation function is excessive. 
To validate the claim, we slightly modified the additional activation function and analyzed the effect of threshold shifting technique.
We replaced the additional PReLU layers with LeakyReLU layers and varied the slope of the negative range of the LeakyReLU layers from 0 to 1.
Note that when the slope is 1, the LeakyReLU layer becomes an identity function and therefore the model becomes the same as the ResNet-20 model without PReLU layers.
\begin{figure} 
    \centering
    \includegraphics[width=0.95\linewidth]{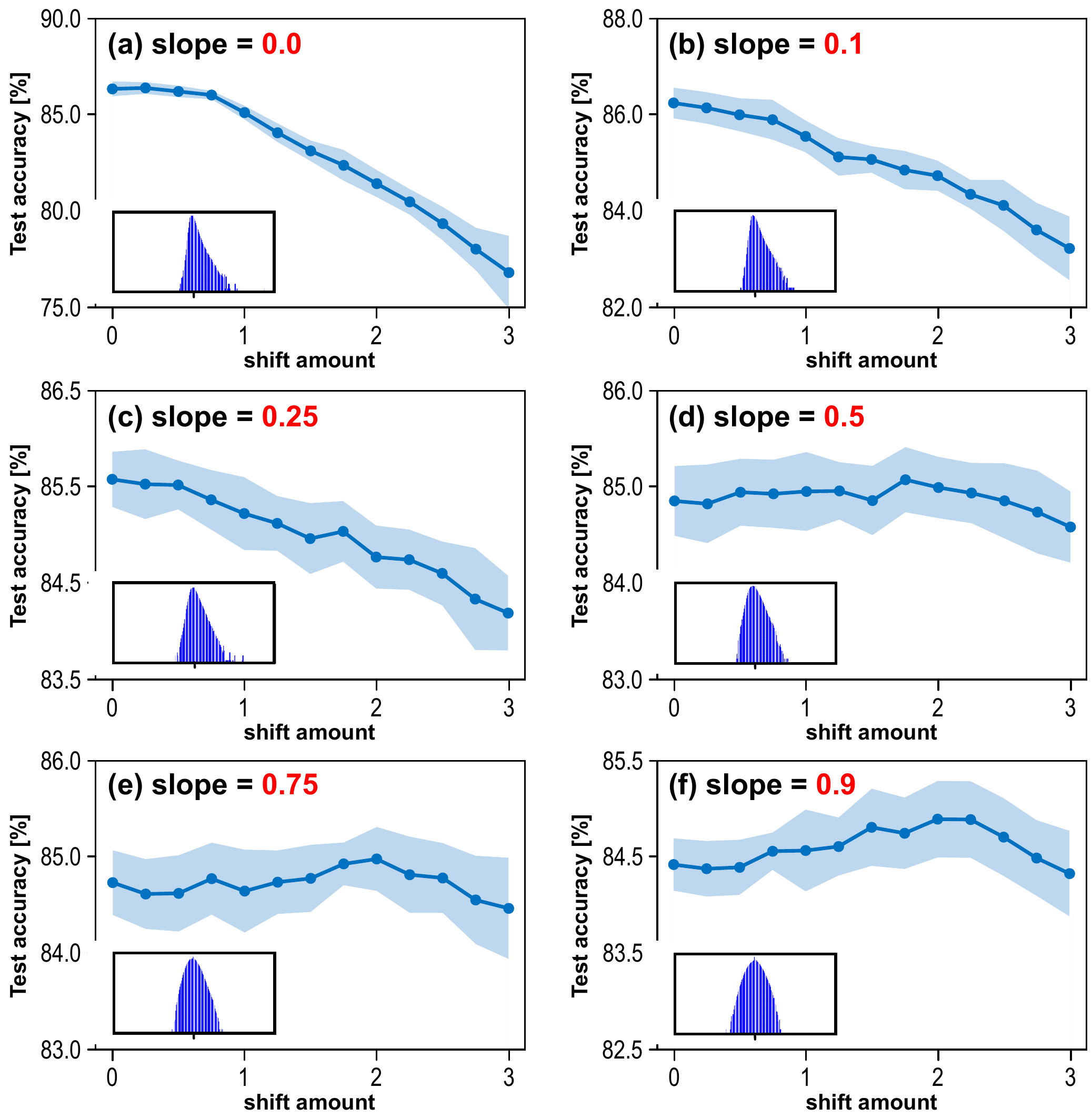}
    \caption{Effect of the slope of the negative range of LeakyReLU layers on the dependence of the test accuracy on the threshold shift amount. The slope of the negative range of LeakyReLU is (a) 0.0, (b) 0.1, (c) 0.25, (d) 0.5, (e) 0.75, and (f) 0.9. 
    Inset: distribution of pre-activation when shift amount is 0 in each case.
    }
    \label{fig:leakyrelu}
    \vspace{-3mm}
\end{figure}
Fig.~\ref{fig:leakyrelu} shows the training results of ResNet-20 with LeakyReLU layers with different slopes.
As the slope of the negative range of LeakyReLU decreases, the distortion of the activation distribution becomes more severe.
When the slope is close to 1, the LeakyReLU layer does not skew the distribution of pre-activation much.
Hence, the distribution of pre-activation values is very close to a Gaussian distribution with zero mean.
As a result, the accuracy improvement by shifting the threshold of binary activation function was clearly observed as in the ResNet-20 model without PReLU layers.
However, as the slope decreases, the LeakyReLU layer makes the activation distribution unbalanced as shown in Fig.~\ref{fig:leakyrelu}.
Note that LeakyReLU layer scales down the negative values only so that the mean value of the distribution increases while the median value of the distribution does not change.
Therefore, it has a similar effect to shifting the threshold of binary activation function in that the activation distribution becomes unbalanced.

\section{Conclusion}
In this paper, we analyzed the impact of activation distribution on the accuracy of BNN models.
While previous BNNs used sign function as binary activation function which balances the distribution of binary activation, we claim that the accuracy of BNN models can be improved when the distribution of binary activation is unbalanced.
By simply shifting the thresholds of binary activation functions, we demonstrated that the accuracy of previous BNN models could be further improved.
We also identified that the unbalanced activation distribution partly accounts for the improved accuracy of BNN models which used additional activation functions. 

\small{
\textbf{Acknowledgement:} This work was in part supported by the National Research Foundation of Korea (NRF) grant funded by the Korea government (MSIT) (NRF-2020R1A2C2004329), Samsung Research Funding Center under Project Number SRFC-TC1603-51, and Institute of Information \& communications Technology Planning \& Evaluation (IITP) grant funded by the Korea government (MSIT) (No.2019-0-01906, Artificial Intelligence Graduate School Program(POSTECH)).
}

{\small
\bibliographystyle{ieee_fullname}
\bibliography{egbib}
}

\section*{S1 Experimental setup}
In this work, we used the vgg-small model as a baseline network for various experiments.
The network consists of 4 convolution layers with 64, 64, 128, 128 output channels and 3 dense layers with 512, 512, and 10 output neurons.
A batch normalization(BN) layer is placed between the convolution layer and the activation layer following the typical BNN block structure.
Max pooling layer was used after each of the latter three convolution layers.

In Sec.3.1, to evaluate the effect of shifting the activation function along the x- or y-axis and that of increasing the range of hardtanh function, we used the vgg-small model with binary weight and full-precision activation.
In Sec.3.2 and 3.4, we used the BNN version of the vgg-small model (binary weight and binary activation).
For binary weight, we used the channel-wise high precision scaling factor following \cite{xnor}.
Neither weight decay nor weight clipping were used in this work.
We trained both the full precision and binary models for 200 epochs using Adam optimizer~\cite{adam} with an initial learning rate of 0.01. 
We used a batch size of 256, and scheduled the training using cosine annealing method~\cite{cosineanneal} with 5 epochs of warmup. 

In Sec.3.2.2, we tested two additional model architectures on MNIST dataset. 
The 2-layer MLP model has two fully connected layer with 512 and 10 output channels each. 
The model structure is as follows: FC1 -  BN1 - BinAct - FC2 - BN2. 
The network is simplified as much as possible in order to rule out any effect other than threshold shifting of the single binary activation. 
The binary LeNet-5 has the network structure adopted from~\cite{lenet5} and has an additional BN layer prior to every binary activation. 
We trained the networks for 30 epochs without warmup. 
All other hyperparameters are set the same as the ones used in the experiments conducted on CIFAR-10.
We also investigated the sensitivity to the optimizer using SGD with momentum.
We used an initial learning rate of 0.1 and momentum of 0.9 in this case.

In Sec.3.3, we evaluated several BNN models for ImageNet dataset.
We followed the training recipe of the original authors for each model.
We only modified the activation function to apply the proposed threshold shifting technique.

In Sec.3.5, we used ResNet-20 model to examine the effect of the additional activation function.
The model architecture is described in Fig.9a.
We used the same optimizer and training recipe as in Sec.3.2.

\section*{S2 BN bias initialization shifting as an alternative to the threshold shifting}
As described in Sec.3.4.1, the threshold of binary activation function can be absorbed to the BN bias term in case the binary activation layer follows the BN layer.
Therefore, the proposed threshold shifting technique can also be implemented by shifting initialization values of the BN bias instead.
It can be easily seen that the threshold value can be merged into the bias term of the previous BN layer by simply subtracting the value from BN bias.
Since the threshold value is fixed during the network optimization, the term can be merged prior to training, namely BN bias initialization. 
Instead of using a positive threshold value for binary activation, we can initialize the BN bias with the negative of that value to produce an unbalanced activation distribution in the network.
Therefore, the proposed threshold shifting technique does not require any additional computation or resource during training and inference.

\end{document}